\documentclass{article}
\usepackage{spconf,amsmath,graphicx,hyperref}

\usepackage{cleveref}
\usepackage{footnote}
\usepackage{booktabs}

\title{HIGH QUALITY UNDERWATER IMAGE COMPRESSION WITH ADAPTIVE COLOR CORRECTION}
\makeatletter
\gdef\@name{%
  {\em Yimin Zhou$^{1,*}$
\thanks{This work is supported in part by the National Natural Science Foundation of China under grant 92467204, 62472249, 62576122, 62301189, and Shenzhen Science and Technology Program under Grant JCYJ20220818101014030, KJZD20240903102300001 and KJZD20240903103702004.}, Yichong Xia$^{1,4,*}$
\thanks{$^*$These authors contributed equally to this work.}, Sicheng Pan$^1$, Bin Chen$^{2,4,\dag}$
\thanks{$^\dag$Corresponding author.}, Yaowei Li$^3$} \\[3pt]  
  {\em Jiawei Li$^3$, Mingyao Hong$^4$, Zhi Wang$^1$, Yaowei Wang$^{2,4}$}  
}
\makeatother

\address{$^1$Tsinghua Shenzhen International Graduate School, Tsinghua University\\
         $^2$Harbin Institute of Technology, Shenzhen\\
         $^3$Huawei Technology\\
         $^4$Peng Cheng Laboratory}

\begin{document}
\ninept
\maketitle
\begin{abstract}

With the increasing exploration and exploitation of the underwater world, underwater images have become a critical medium for human interaction with marine environments, driving extensive research into their efficient transmission and storage. However, contemporary underwater image compression algorithms fail to adequately address the impact of water refraction and scattering on light waves, which not only elevate training complexity but also result in suboptimal compression performance. To tackle this limitation, we propose High Quality Underwater Image Compression (HQUIC), a novel framework designed to handle the unique illumination conditions and color shifts inherent in underwater images, thereby achieving superior compression performance. HQUIC first incorporates an Adaptive Lighting  and Tone Correction (ALTC) module to adaptively predict the attenuation coefficients and global light information of images, effectively alleviating issues stemming from variations in illumination and tone across underwater images. Secondly, it dynamically weights multi-scale frequency components, prioritizing information critical to distortion quality while discarding redundant details. Furthermore, we introduce a tone adjustment loss to enable the model to better balance discrepancies among different color channels. Comprehensive evaluations on diverse underwater datasets validate that HQUIC outperforms state-of-the-art compression methods, demonstrating its effectiveness.

\end{abstract}
\begin{keywords}
Underwater Image, Image Compression 
\end{keywords}

\section{Introduction}
Underwater imaging stands as a crucial means of understanding the marine world and serves as an essential carrier for deep-sea exploration missions and underwater operations, thereby garnering significant attention. However, the complex and dynamic underwater environment, coupled with signal attenuation and distortion, imposes an upper bound on the data transmission capacity between underwater acquisition robots and ground stations. This limited transmission capability is mismatched with the ever-growing demands for underwater exploration and underwater imagery, thus necessitating the development of efficient compression algorithms.


\begin{figure}[t!]
  \includegraphics[width=0.95\linewidth]{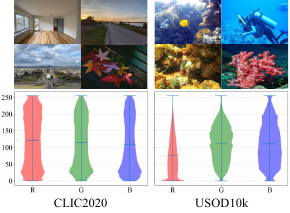}
  \caption{The three channel intensity distribution across various datasets, including the terrestrial image dataset CLIC2020 and the underwater image dataset USOD10k.}
  \label{pic:violin}
\end{figure}

\begin{figure*}[t!]
  \centering
  \includegraphics[width=\textwidth]{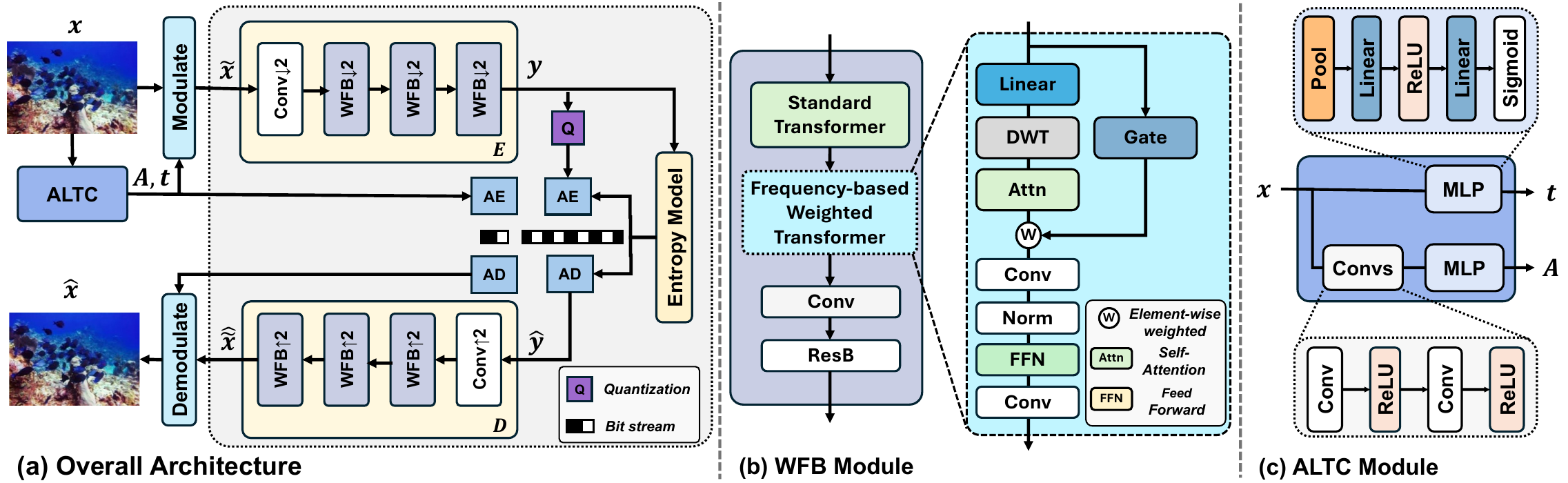}
  \caption{Overview of the proposed methods. (a) presents the overall architecture of the pipeline, and (b), (c) shows the detailed architecture of the WFB module and ALTC module respectively.   }
  \label{pic:pipeline}
\end{figure*}
Unlike the compression of terrestrial images, underwater images exhibit unique characteristics, which hinders their efficient compression. 
Underwater images typically display a cyan-green tint caused by light scattering and diffuse reflection in water, as well as the ubiquitous presence of floating flocculents in the ocean. This results in significant uneven color shifts in the images. To investigate this, we analyzed three RGB channel values of images from commonly used datasets, including the terrestrial image dataset CLIC2020 \cite{toderici2020workshop}, as well as the underwater image dataset USOD10k \cite{hong2023usod10k}. As shown in \cref{pic:violin}, underwater image datasets exhibit significantly smaller red channel values and larger blue and green channel values compared to terrestrial image datasets. This indicates that the channel redundancy in underwater images differs markedly from that in terrestrial images. This indicates that underwater images differ significantly from natural images in terms of color distribution.

In recent years, various methods for underwater image compression have been proposed. However, these approaches have not fully eliminated the spatial and channel redundancies inherent in underwater images, resulting in suboptimal performance. These methods can be broadly categorized into two types.
The first category involves traditional methods for underwater image compression \cite{li2010low, he2020low,zhang2015seafloor}. These methods utilize dimensionality reduction and mathematical analysis to extract the primary features of underwater images. 
The second category comprises deep learning-based methods \cite{krishnaraj2020deep,li2022extreme, fang2023priors,chen2024semantic,xia2026towards,xiadiffpc,qin2024progressive}. These approaches leverage data-driven training to automatically extract features, achieving better performance. However, traditional methods largely rely on manually constructed feature extraction processes, and models based on convolutional networks lack the capability to capture long-range information. Furthermore, due to their failure to address the illumination characteristics of underwater images, they still yield suboptimal results.

To further enhance the performance, we propose a novel underwater image compression framework, HQUIC. 
Firstly, HQUIC incorporates the Adaptive Lightning and Tone Correction Module (ALTC), which enables the model to take into account the depth and lighting conditions of different underwater areas and adaptively predict the attenuation coefficients of different channels and global light information.
Moreover, HQUIC designs a tone adjustment loss, which effectively improves the model's compression performance on underwater images by balancing the intensity of different color channels.  As a result, it reduces the difficulty of model training caused by lighting and tone differences among underwater images. 
Finally, HQUIC utilizes the Frequency-based Weighted Transformer Module (FBWT) to process and weight information in different frequency domains at multiple scales, thus enhancing the compression performance of the model.
Experiments demonstrate that our method outperforms all baselines on different underwater image datasets and achieves state-of-the-art results.

 Our contributions are summarized as follows:
 \begin{itemize}
     \item We propose  a novel underwater image compression model, HQUIC, which achieves state-of-the-art performance compared with traditional compression methods and learned lossy image compression methods.
     \item We propose the ALTC module, which effectively mitigates the performance loss caused by lighting and tone. Furthermore, a tone adjustment loss is proposed to balance the intensity of different color channels.
     \item In addition, we put forward the FBWT module to assess the significance of the frequency domain, thereby enhancing the compression quality.
 \end{itemize}

\section{Methods}


\subsection{Adaptive Lighting and Tone Correction}
\label{sec:ALTC}
Underwater images are significantly influenced by water quality and lighting conditions, resulting in a pronounced blue-green color cast. To model and solve the special condition,
the Jaffe-McGlamery model \cite{jaffe1990computer} is one of the most widely utilized models in underwater image processing, which can be simplified as follows:
\begin{align}
\label{eq:3}
I_i(a,b) = J_i(a,b)t_i(a,b) + A_i(1-t_i(a,b))
\end{align}
where $J(a,b)$ is the standard image,  $A$ is background light and $t(a,b)$ is the refractive index. Subscript $i$ represents the color channel index which includes red, green and blue.

Motivatied by the Jaffe-McGlamery imaging model, our neural network adaptively predicts global illumination conditions and per-channel attenuation coefficients. Specifically, we integrate a lightweight correction module prior to the compression pipeline. This module processes input underwater images $x$ through a global illumination estimation network $E_A$ and a refractive index prediction network $E_t$ to get global illumination conditions $A$ and per-channel attenuation coefficients $t$, and we correct the original input image $x$ to $\tilde{x}$. This operation can be expressed as
\begin{align}
\tilde{x} &= \frac{x-E_A(x)(1- E_t(x))}{ E_t(x)}
\end{align}

\begin{figure*}[t!]
  \centering
  \includegraphics[width=0.95\textwidth]{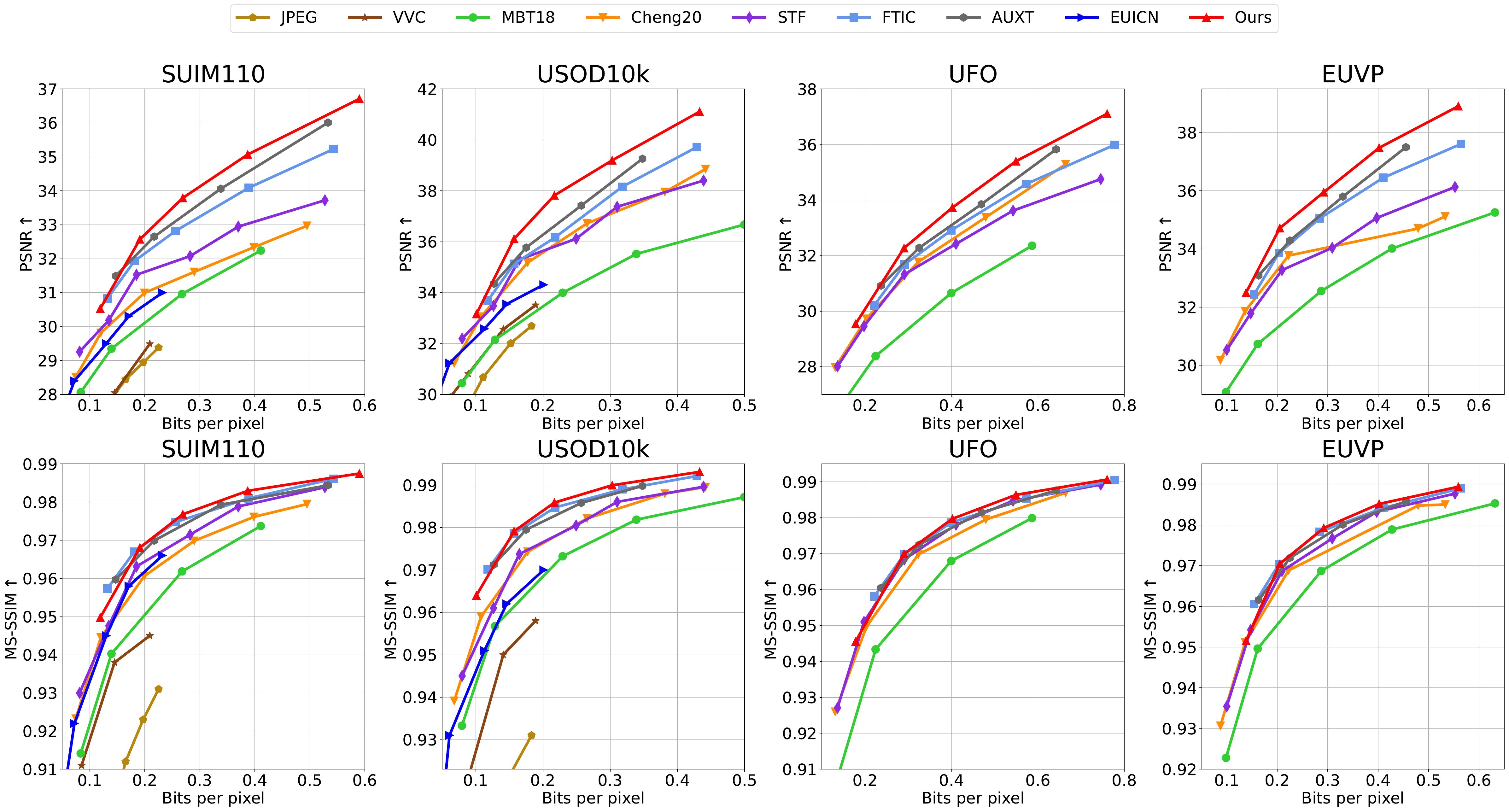}
  \caption{RD-performance between bitrates and metrics, including PSNR and MS-SSIM in four different underwater image datasets. The $\uparrow$ means higher is better.}
  \label{pic:rdcurve}
\end{figure*}
After transmission, we get quantized $\hat{\tilde x}$, and we need to supplement the original global illumination information and underwater refractive index information into the reconstructed image, so that the reconstructed image preserves as much of the original underwater image's information as possible. The specific reconstruction method is as follows:
\begin{align}
\hat{x} &= E_A(x)(1-E_t(x))+\hat{\tilde{x}}\cdot E_t(x)
\end{align}
where $\hat{y}$ is the reconstructed corrected image from the compression model, and $E_A(x)$ and $E_t(x)$ are losslessly transmitted to the decoder side.


\subsection{Tone Adjustment Loss}
To better balance the tone bias of the input image, we propose a tone adjustment loss. 
First, we define $RG$ and $YB$ to reflect the red-green difference and yellow-blue difference of the modulated image $\tilde{x}$, respectively. Then, we sort all $K$ pixels of the image in ascending order according to their corresponding $RG$ or $YB$ values, and use the parameters $\alpha_L$ and $\alpha_R$ to control the proportion of selected pixels. Based on the selected pixels, we calculate the average value between the upper and lower bounds, which reflects the degree of color bias in the image. The specific steps are as follows:
\begin{align}
\label{eq:uicm}
    RG,YB&=(R-G), (R+G)/2-B,  \\
\mu_{channel}&=\sum_{i=\alpha_LK}^{(1-\alpha_R)K}\frac{channel_{i}}{(1-\alpha_L-\alpha_R)K},\\
L_{TA}(\tilde x)&=\sqrt{\mu_{RG}(\tilde x)^2+\mu_{YB}(\tilde x)^2}.
\end{align}
In summary, the training loss of the model is as follows:
\begin{align}
L=R(\hat y)+\lambda D(x,\hat x)+\beta L_{TA}(\tilde x),
\end{align}
where $R$ represents the rate and $D$ represents the distortion. Following the previous work, MSE is utilized as $D$.

\subsection{Frequency-based Weighted Transformer Module}
\label{sec:fwt}
In order to better decouple the information of underwater images and achieve a more compact representation, we adopt a Frequency-based Weighted Transformer (FBWT) Module  to analyze the importance of the frequency domain of features at multiple levels, as shown in \cref{pic:pipeline} (b). 

Specifically, the FBWT is composed of a gating network $G$, a Discrete Wavelet Transform $DWT$, and four expert Attention networks $Attn_i (i=1,2,3,4)$, where $i$ represents the four sub-bands. The gating network receives the features $f^j$ of a certain intermediate layer $j$ and outputs the weighting coefficients of four sub-bands. The features $f^j$ are also input into the $DWT$ simultaneously, which decomposes the features into four sub-bands, namely HH, HL, LH, and LL, and sends the four types of sub-band features into the corresponding expert Attention networks $Attn_i$, respectively. Finally, the processing results of the four experts are weighted. The specific steps are described as follows:
\begin{align}
    v^j &= Merge(G(f^j)_i\cdot Attn_i(DWT(f^j)_i))
\end{align}
Among them,  $Merge$ represents the channel-level merge operations. Then $v^j$ can be sent to the following operation.

\section{Experiments}
\begin{figure*}[t!]
  \centering
  \includegraphics[width=\textwidth]{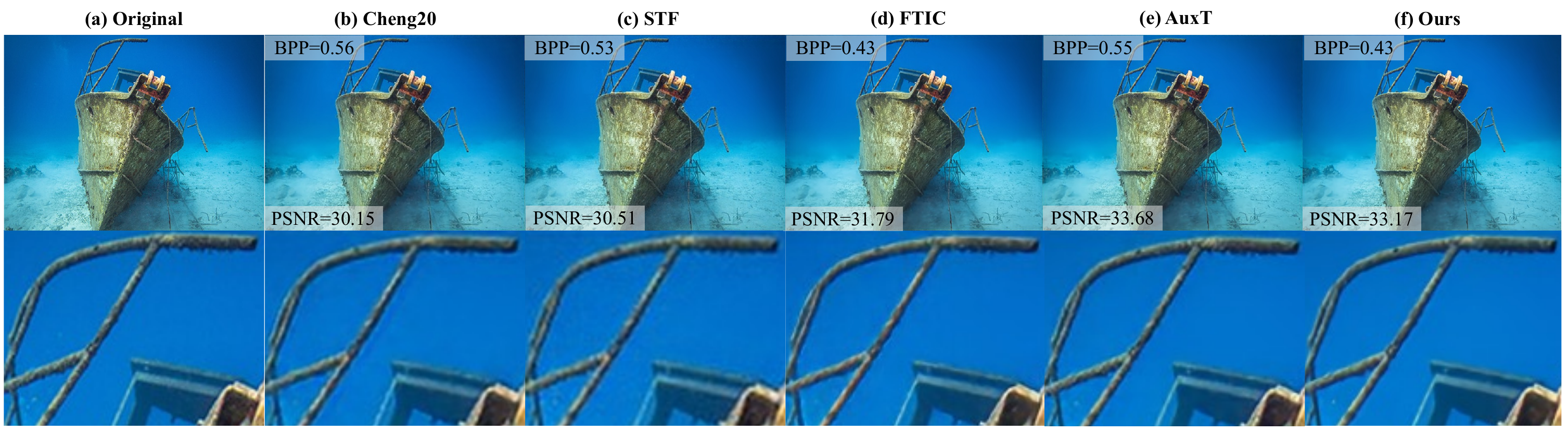}
  \caption{The original images and the decompressed image of different baselines and our method. For each image, the upper corner is labeled as bpp and bottom-left corner is labeled as PSNR. The second row is the zoom-in images of the first row.}
  \label{pic:visual}
\end{figure*}

\begin{table}[t!]
\caption{The BD-rate performance on the USOD and SUIM datasets. Lower is Better.}
\label{tab:bd-main}
\begin{tabular}{c|cccc}
\toprule
 & \multicolumn{2}{c}{USOD} & \multicolumn{2}{c}{SUIM}   \\
 & PSNR      & MS-SSIM      & PSNR      & MS-SSIM          \\
 \midrule
 Cheng20&    $-39\%$  &   $-23.0\%$    &    $-24\%$   &  $-21.8\%$                \\
 STF&   $-41\%$  &    $-20.1\%$  &  $-35\%$    &  $-25.6\% $                  \\
 FTIC&   $-47\%$     &   $-43.7\% $ &   $-50\% $  &    $    -43.5\%  $ \\
 AuxT& $-52\%$ & $-42.0\%$ & $-54\%$ & $-39.0\%$\\
 Ours&   $\boldsymbol{-56\%} $    &   $\boldsymbol{-44.9\% }$ &   $ \boldsymbol{-54\%}  $  &    $  \boldsymbol{-44.0\%} $  \\
 
\bottomrule
\end{tabular}
\end{table}
\subsection{Experimental Settings}
\textbf{Training Detail.}
We conducted the training on the USOD10k dataset, which contains 9229 underwater images  of different sizes, and we randomly crop it into $256\times256$. The implementations are based on the PyTorch \cite{paszke2019pytorch} and the CompressAI library \cite{begaint2020compressai}, where a single NVIDIA GeForce RTX 3090 GPU with 24GB memory was utilized in training. We employed the Adam optimizer with a learning rate of 1e-4.  We trained our models for 200 epochs with a batch size of 8. In order to achieve various compression ratios, the parameter for controlling the compression ratio $\lambda$ is set as \{$0.0025$, $0.0067$, $0.013$, $0.025$, $0.0483$\}. Hyperparameter $\beta$ is set as $0.1$.

\textbf{Evaluation.}
We tested our method on multiple datasets, including the USOD10k test dataset \cite{hong2023usod10k}, the SUIM dataset \cite{islam2020semantic}, the UFO dataset \cite{islam2020simultaneous}, and the EUVP dataset \cite{islam2020fast},
where both high-quality and low-quality images are included. We use PSNR and MS-SSIM metrics to measure the performance.

\textbf{Baselines.}
 The baseline methods included traditional compression methods like JPEG \cite{taubman2002jpeg2000} and VVC, and the widely used learned image compression methods MBT18 \cite{minnen2018joint}, Cheng20 \cite{cheng2020learned}, STF \cite{zou2022devil}, FTIC \cite{lifrequency}, AuxT \cite{lidisentangled2025}, and the underwater image compression method EUICN \cite{li2024euicn}.
To ensure the fairness of the comparison, we used their official code and retrained all the baselines on the same underwater dataset. For EUICN, since the original paper does not provide open-source code, the data used here is directly taken from the original paper.

\subsection{Rate-Distortion Performance}
The comparison results of our proposed method with the baselines in PSNR and MS-SSIM in underwater image datasets are shown in the \cref{pic:rdcurve}.
As depicted in the \cref{pic:rdcurve}, our proposed method surpasses various existing approaches, which demonstrates that our method achieves excellent performance across all underwater image datasets. Although some methods designed for terrestrial images exhibit poor performance when dealing with underwater image scenes, even after being retrained on underwater images, the method we proposed consistently maintains excellent underwater image compression performance in PSNR, especially at high bitrates. This is because at high bitrates, the method we proposed can effectively address the reconstruction difficulties caused by the color difference issues existing in underwater environments. 

Moreover, we use the BD-rate metric to present the average performance gain. As can be seen from the table, our method saves the most $56\%$ and $54\%$ bit-rate to achieve the same PSNR, and $45\%$ and $44\%$ bit-rate to achieve the same MS-SSIM.

In \cref{pic:visual}, we provide the visual results. In order to better visualize the differences , we have enlarged the local areas shown in the bottom rows. Despite the interference from underwater conditions, our proposed method reconstructs the details of the images better. 

\subsection{Ablation Study}
We conduct ablative experiments on modules we proposed, as shown in \cref{tab:bd-ablation}. We utilized BD-rate as a metric to gauge the extent of increase in bitrate at the same level of distortion compared to the reference point. $\boldsymbol{0\%}$ means the full model which is set as the reference point, and other positive values represent the performance degradation caused by the absence of corresponding modules.

Among them, the ATLC module significantly enhanced the PSNR metric of the reconstructed images.  Without this module, on average, nearly 30\% additional bits are required to achieve the same PSNR. By accurately compensating for the unique lighting conditions in underwater scenes, the ATLC module reduced the noise and distortion in the reconstructed images, leading to a substantial improvement in the PSNR value. Due to the fact that FBWT effectively decomposes images in the frequency domain, it enables the model to better process information from various directions. As a result, there are considerable $7.4\%$ and $11.57\%$ improvements in both the PSNR and MS-SSIM metrics in the USOD dataset.

\begin{table}[t!]
\caption{The BD-rate performance on the USOD and SUIM datasets. The full model is selected as the reference point.}
\label{tab:bd-ablation}
\begin{tabular}{c|cccc}
\toprule
 & \multicolumn{2}{c}{USOD} & \multicolumn{2}{c}{SUIM}   \\
 & PSNR      & MS-SSIM      & PSNR      & MS-SSIM          \\
 \midrule
 w/o ATLC&   $ 29.3\%$  &  $ 4.31\%$    &  $ 22.8\% $  & $ 1.64\%$                \\
 w/o FBWT&    $ 7.4\%$     &     $   11.57\%$      & $5.9\%$    &    $6.14\%$   \\
 Full model&      $\boldsymbol{0\%}$  &  $\boldsymbol{0\%}$ & $\boldsymbol{0\%}$     & $\boldsymbol{0\%}$   \\
 
\bottomrule
\end{tabular}
\end{table}
\section{Conclusion}
In conclusion, HQUIC presents a novel method specifically designed for the illumination and color characteristics of underwater images. By introducing the ALTC module and tone adjustment loss, color shifts and illumination effects are mitigated. Additionally, a frequency-based weighted attention mechanism dynamically prioritizes multi-scale frequency components critical for perceptual quality. Extensive experiments on diverse underwater datasets demonstrate that HQUIC achieves state-of-the-art rate-distortion performance, substantially outperforming both traditional compression methods and terrestrial image compression models fine-tuned for underwater scenarios.

\bibliographystyle{IEEEbib}
\bibliography{strings,refs}

\end{document}